\documentclass[format=sigconf]{acmart}
\AtBeginDocument{%
  }

\acmISBN{978-1-4503-XXXX-X/2018/06}

\begin{document}

\title{Evaluating Real-World Generalizability of Algorithm Selection Models}


\author{Gjorgjina Cenikj}
\affiliation{
  \institution{Computer Systems Department\\ Jo\v{z}ef Stefan Institute}
  \city{Ljubljana} 
  \country{Slovenia}
}
\email{gjorgjina.cenikj@ijs.si}

\author{Jakub Kudela}
\affiliation{
  \institution{ Brno University of Technology}
  \city{Brno} 
  \country{Czech Republic}
}
\email{Jakub.Kudela@vutbr.cz}
\author{Eva Tuba}
\affiliation{
  \institution{Computer Systems Department\\ Jo\v{z}ef Stefan Institute}
  \city{Ljubljana} 
  \country{Slovenia}
}
\email{eva.tuba@ijs.si}

\author{Tome Eftimov}
\affiliation{
  \institution{Computer Systems Department\\ Jo\v{z}ef Stefan Institute}
  \city{Ljubljana} 
  \country{Slovenia}
}
\email{tome.eftimov@ijs.si}

\renewcommand{\shortauthors}{Cenikj et al.}

\begin{abstract}
Algorithm Selection (AS) aims to automatically identify the most suitable optimization algorithm for a given problem instance by leveraging measurable problem characteristics and historical performance data. In this study, we investigate the generalization ability of AS models across both synthetic and real-world optimization landscapes. We consider two widely used academic benchmark suites (BBOB and CEC) and two real-world problem sets (robotics trajectory optimization tasks and unmanned aerial vehicle path-planning problems). Through a systematic cross-benchmark evaluation, we analyze how AS models transfer between domains, identify where generalization succeeds or breaks down, and highlight the challenges that arise when applying AS in realistic, domain-specific contexts. Our findings provide insights into the robustness of current AS approaches and inform the development of more reliable, broadly applicable AS systems for real-world optimization.
\end{abstract}





\maketitle

\section{Introduction}
Algorithm Selection (AS)~\cite{as_survey} is the task of automatically identifying the most suitable optimization algorithm for solving a given problem instance, based on measurable characteristics of the problem (e.g., landscape features) and past performance data. Rather than relying on trial-and-error or expert intuition, AS systems aim to leverage data-driven models to make informed algorithm choices. AS has become increasingly relevant as the number and diversity of optimization algorithms continue to grow, and no single algorithm can consistently outperform others across all problem types.

While existing benchmark suites such as the Black-Box Optimization Benchmarking (BBOB)~\cite{bbob} and the benchmarks provided by the IEEE Congress on Evolutionary Computation (CEC) Special Sessions and Competitions on Real-Parameter Single-Objective Optimization~\cite{cec2013, cec2014, cec2015, cec2017} series have been essential for developing and evaluating AS models in controlled settings, they cover only a limited portion of the vast landscape of real-world optimization problems. These synthetic benchmarks are valuable for systematic comparisons because they contain well-defined problem classes with controlled properties. However, they often lack the structural complexity, noise patterns, heterogeneity, and domain-specific constraints that characterize real-world problems. For instance, an analysis of the BBOB benchmark and real-world crashworthiness problems~\cite{long2022learning} shows that in the problem landscape space, crashworthiness problems are separated and different from the BBOB. Similar findings were observed for unmanned aerial vehicle (UAV) problems~\cite{jakub_uav_problems}.
As a result, AS models that perform well on standard benchmarks may overfit to synthetic landscapes and fail to generalize when applied to practical tasks.

To address this limitation, in this study we evaluate the generalization ability of AS models across both synthetic and real-world benchmarks. Specifically, we consider two synthetic suites (BBOB and CEC) and two real-world problem sets: a collection of robotics trajectory optimization problems used in~\cite{jakub_robot_problems}, and a set of UAV path planning problems presented in~\cite{jakub_uav_problems}. By systematically testing AS models on this combination of benchmarks, we aim to assess how well they generalize beyond controlled synthetic settings and to identify the challenges that arise when applying AS in realistic, domain-specific contexts. 

\noindent\textbf{Outline:} The remainder of this paper is organized as follows. Section~\ref{sec:related} reviews related work. Section~\ref{sec:methdology} describes the methodology, Section~\ref{sec:results} presents and discusses the results. Section~\ref{sec:conclusion} concludes the paper and outlines directions for future work.

\noindent\textbf{Reproducibility:} The code is available at \url{https://anonymous.4open.science/r/Algorithm_Selection_real_world_generalizability-3DE8/}. The data used for the experiments is available at \url{https://zenodo.org/records/18350950}.

\section{Related Work}
\label{sec:related}
The primary objective of algorithm selection (AS) is to train a machine learning model capable of recommending the most suitable optimization algorithm for previously unseen problems with different landscape characteristics. Achieving this level of generalization, however, remains a significant challenge.

It has been shown in~\cite{urban_transfer_learning} that an AS model based on ELA features, when trained on randomly generated functions~\cite{random_function_generator_matlab}, fails to generalize effectively to the BBOB benchmark. Similar findings have been reported across multiple studies on different benchmarks, consistently highlighting the difficulty of transferring AS models beyond their training distributions.

Generalization across four different benchmarks has been investigated in~\cite{gina_cross_benchmark_generalization}, where the experimental setup involved training an AS model on one benchmark and evaluating it on a completely different one. This work also examined the complementarity between ELA and transformer-based features trained for the task of BBOB problem classification~\cite{transopt}. The results showed that when the training and testing benchmarks have a similar distribution of algorithm performance and share the same single-best solver, feature-based AS models do not outperform a simple baseline that predicts the mean performance of the single best solver on the training benchmark. Additionally, the study pointed out that there are problem instances for which similar landscape features do not guarantee similar algorithm performance, indicating that the examined features lack sufficient discriminatory power to reliably capture algorithm behavior.

The generalization of an AS model trained on the original BBOB problem instances and evaluated on their affine combinations was analyzed in~\cite{gasper_affine_generalization}. The results indicate that while the model can predict algorithm performance well for problems similar to those seen during training, its performance decreases substantially as the evaluation problems deviate from the training distribution, eventually becoming comparable to a baseline that predicts the mean algorithm rank.

A methodology for evaluating how well predictive models trained on one benchmark generalize to another has been proposed in~\cite{assessing_generalizability}. The findings show that patterns of generalizability observed in problem landscape features are closely reflected in algorithm performance outcomes, underscoring the importance of assessing AS models across diverse benchmarks.

In~\cite{gina_hit_the_wall}, a comprehensive analysis of the performance of AS models using ELA features and several recently proposed features based on deep learning~\cite{doe2vec,deepela,transoptas} and topological landscape analysis~\cite{tinytla} has been conducted on the affine combinations of BBOB problems~\cite{MABBOBteloversion}. The results indicate that the ELA features provide the best results in easier evaluation settings, however, none of the investigated feature groups generalize well to unseen problems.

All of the previously mentioned works evaluate the generalization of AS models using academic benchmarks or artificial problem generators. In contrast, in this work, we aim to evaluate the generalization of AS models between academic benchmarks and real-world problems.

\section{Methodology}
\label{sec:methdology}
In this section, we outline our methodology. We first present the problem and algorithm portfolios, then define the metric used to evaluate algorithm performance. We subsequently detail the feature computation process, the training and evaluation procedure for the AS model.

\subsection{Problem Portfolio}

We evaluate algorithm selection performance across four benchmark sources, covering both synthetic and real-world optimization tasks: \textbf{BBOB} -- we use the first 20 instances of each of the 24 problem classes from the Black-Box Optimization Benchmarking suite, using the IOHprofiler \cite{IOHprofiler} implementation; \textbf{CEC} --  we use 15 instances for each of the 47 problems from the CEC real-parameter single-objective benchmark series, taken from \cite{stripinis2024benchmarking}; \textbf{ROB} -- a set of robotics trajectory optimization problems, with 4 settings for the number of positions to reach (2, 4, 6, and 8) and with 100 instances (randomized positions of the points) for each setting, taken from \cite{jakub_robot_problems}; and \textbf{UAV} -- a collection of unmanned aerial vehicle (UAV) path-planning problems, 
    with 4 settings for the number of obstacles (5, 10, 15, and 20) and with 100 instances (randomized layout of the obstacles and environmental conditions) for each setting, taken from \cite{jakub_uav_problems}. The UAV and ROB benchmarks have their natural dimensions as multiples of 3 and 6, respectively. Therefore, our experiments were done in dimensions 6, 12, 18, 24, and 30 for all four benchmarks, which was the main reason for selecting \cite{IOHprofiler} as the implementation of the BBOB suite and \cite{stripinis2024benchmarking} as the implementation of the CEC problems, as both facilitated running the experiments in these ``nonstandard'' dimensions. 

%\todo{@jakub if you can describe UAV and ROB problems better, and how instances are generated}
\subsection{Algorithm Portfolio}

Our algorithm portfolio comprises eight population-based single-objective optimization methods commonly used in continuous black-box optimization. It includes both the best-performing methods of CEC Competitions and ``standard'' methods: Adaptive Gaining-Sharing Knowledge Algorithm (\texttt{AGSK})~\cite{alg_agsk}, Gaining-Sharing Knowledge Based Algorithm with Adaptive Parameters Hybrid with IMODE (\texttt{APGSK-IMODE})~\cite{alg_apgsk_imode}, Differential Evolution (\texttt{DE})~\cite{de}, Four Evolutionary Algorithms with Eigen Crossover (\texttt{EA4eig})~\cite{alg_ea4eig}, Effective Butterfly Optimizer with Covariance Matrix Adapted Retreat (\texttt{EBOwithCMAR})~\cite{alg_ebo}, Enhanced ELSHADE\_SPACMA Algorithm\\(\texttt{ELSHADE\_SPACMA})~\cite{alg_lshade_spacma}, Linear Population Size Reduction SHADE (\texttt{LSHADE})~\cite{alg_lshade}, and  Particle Swarm Optimization (\texttt{PSO})~\cite{pso}.

We use the same MATLAB implementations and parameter settings\footnote{\url{https://github.com/JakubKudela89/Benchmarking_Black_Box_Optimization_Algorithms}} of the methods that were used in \cite{stripinis2024benchmarking}. The \texttt{AGSK}, \texttt{APGSK-IMODE}, \texttt{EA4eig}, \texttt{EBOwithCMAR}, \texttt{ELSHADE\_SPACMA}, and \texttt{LSHADE} methods were selected based on their complementary performance reported in \cite{stripinis2024benchmarking}, where each was among the top methods for problems for various benchmark sets with different high-level features, dimensions, and available computation budgets. We performed 10 independent runs of all the algorithms on each problem/setting-instance-dimension triplet for the four benchmarks. As both the UAV and ROB benchmarks are variable dimension problems with high computational costs for the evaluations, the evaluation budgets for our experiments were set to $\lfloor6\cdot20000/d\rfloor$ (i.e., 20000 for $6d$ and 4000 for $30d$), in a fashion similar to experiments performed in \cite{jakub_robot_problems}.
% \todo{@jakub: add rationale for algorithm selection (complementarity), implementation details, parameter settings, and evaluation budget specification across dimensions.}

%\todo{@jakub, if you can add: how algorithms were selected for complementarity, which library is used to run algorithms, how parameters are set, how budget is set for different dimensions}

\subsection{Algorithm Performance Metric}

\label{subsec:algorithm_performance_metric}

To evaluate the algorithms, we adopt a fixed-budget setup, where the assessment focuses on the best objective function value achieved after a predetermined number of iterations. The algorithm’s performance is measured using the normalized precision metric proposed in~\citep{gina_cross_benchmark_generalization}, defined as follows.

Let $y_{arp}$ denote the objective function value corresponding to the best solution obtained by algorithm $a$ during run $r$ for problem instance $p$. This value is normalized relative to the range of solutions discovered by all other algorithms for the same problem instance and initial population (i.e., within the same run). Denoting by $b_{rp}$ and $w_{rp}$ the best and worst objective values found across all algorithms for problem instance $p$ in run $r$, we define the scaled best value achieved by algorithm $a$ as:

\begin{equation}
s_{arp} = \frac{y_{arp} - b_{rp}}{w_{rp} - b_{rp}}.
\label{eq:algorithm_performance_metric}
\end{equation}

This measure reflects how well the best solution found by algorithm $a$ in run $r$ compares to the other algorithms executed with the same initial conditions. The normalized precision score of algorithm $a$ for problem instance $p$ is then computed as the mean of $s_{arp}$ over all runs.

\subsection{Feature Calculation}

We derive ELA features from the following groups using the \texttt{flacco} Python package~\cite{pflacco}: \texttt{disp}, \texttt{ela\_distr}, \texttt{ela\_level}, \texttt{ela\_meta}, \texttt{ic}, \texttt{nbc}, and \texttt{pca}. In total, this results in 62 features per problem instance. These specific groups were chosen because they do not require extra function evaluations.

Previous studies~\cite{ela_y_normalization, gina_ela_generalization_scaling} have demonstrated that scaling the objective function values to the range [0,1] using min–max normalization can enhance the generalization of ELA features. Since the problem benchmarks being used have decision variables in different ranges, we explore the following preprocessing steps for calculating the ELA features:
\begin{itemize}
    \item None - the decision variables and objective function values are left within their original ranges
    \item y-scaling - only the objective function values are scaled within the range [0,1] using min-max scaling
    \item x-y-scaling - both the decision variables and the objective function values are scaled within the range [0,1] using min-max scaling
\end{itemize}
Additionally, we explore two sample sizes for calculating the features
\begin{itemize}
    \item large sample - 10,000 samples per problem, regardless of dimension
    \item small sample - We downsample the large sample by randomly sampling 50$d$ of the original samples, since this is a commonly used sample size~\cite{pascal_low_budget_ela,seiler2024learned, gina_hit_the_wall}.
\end{itemize}

\subsection{Automated Algorithm Performance Prediction Model}

We adopt a Random Forest (RF) model to perform multi-target regression, where the objective is to simultaneously predict the performance score of each algorithm according to Equation~\ref{eq:algorithm_performance_metric}. The RF model is chosen because of its strong performance on tabular datasets~\cite{shwartz2022tabular}, its robustness to heterogeneous feature scales, and the interpretability offered by its feature importance scores. We use the default settings of the \texttt{scikit-learn}~\cite{scikit-learn} implementation (version 1.2.2), including 100 estimators, the Gini impurity split criterion, \texttt{min\_samples\_split}=2, \texttt{min\_samples\_leaf}=1, \texttt{min\_impurity\_decrease}=0, and \texttt{max\_features} set to the square root of the total number of features; all other parameters are kept at their default value. Separate RF models are trained for each feature representation. Model performance is compared against a simple ``dummy’’ baseline that predicts the mean algorithm performance observed in the training set.

\subsection{Algorithm Selector Performance Metric}

The quality of the algorithm selector is assessed by considering the true performance of the algorithm that the ML model predicts to be the best for each problem instance. Specifically, the algorithm selection performance is defined as:

\begin{equation}
AS\_performance = \frac{1}{|\mathcal{P}|} \sum_{p \in \mathcal{P}} \left[ 1 - (s_{sp} - s_{bp}) \right]
\label{eq:AS_metric}
\end{equation}

Here, $s_{sp}$ denotes the predicted-best algorithm’s score on problem instance $p$, while $s_{bp}$ is the score of the true best-performing algorithm for that instance. The metric ranges from 0 to 1: a score of 0 corresponds to consistently selecting the worst algorithm, whereas a score of 1 indicates perfect prediction of the best algorithm across all problem instances.

\subsection{Model Training and Evaluation}
Most of our experiments target a cross-benchmark evaluation of the AS models, meaning that we train the AS model on one of the benchmarks, and test on all four benchmarks. We perform the model training 5 times on each benchmark, each time splitting the benchmark used for training into a training and testing set. In the case of BBOB and CEC, the training and testing set do not contain instances of the same problem (all instances from one problem go either in the training or testing set). We train separate models for each problem dimension, each type of feature preprocessing and each benchmark.
We also perform an experiment where we merge the benchmarks together (using a single instance from BBOB and CEC, since instances of the same problem are similar to each other).
\section{Results}
\label{sec:results}
We start by presenting the raw performance data obtained by running the algorithm portfolio on the selected benchmarks, followed by the results of the algorithm selection.

\subsection{Algorithm Performance Results}
\begin{figure}
    \centering
    \includegraphics[width=\linewidth]{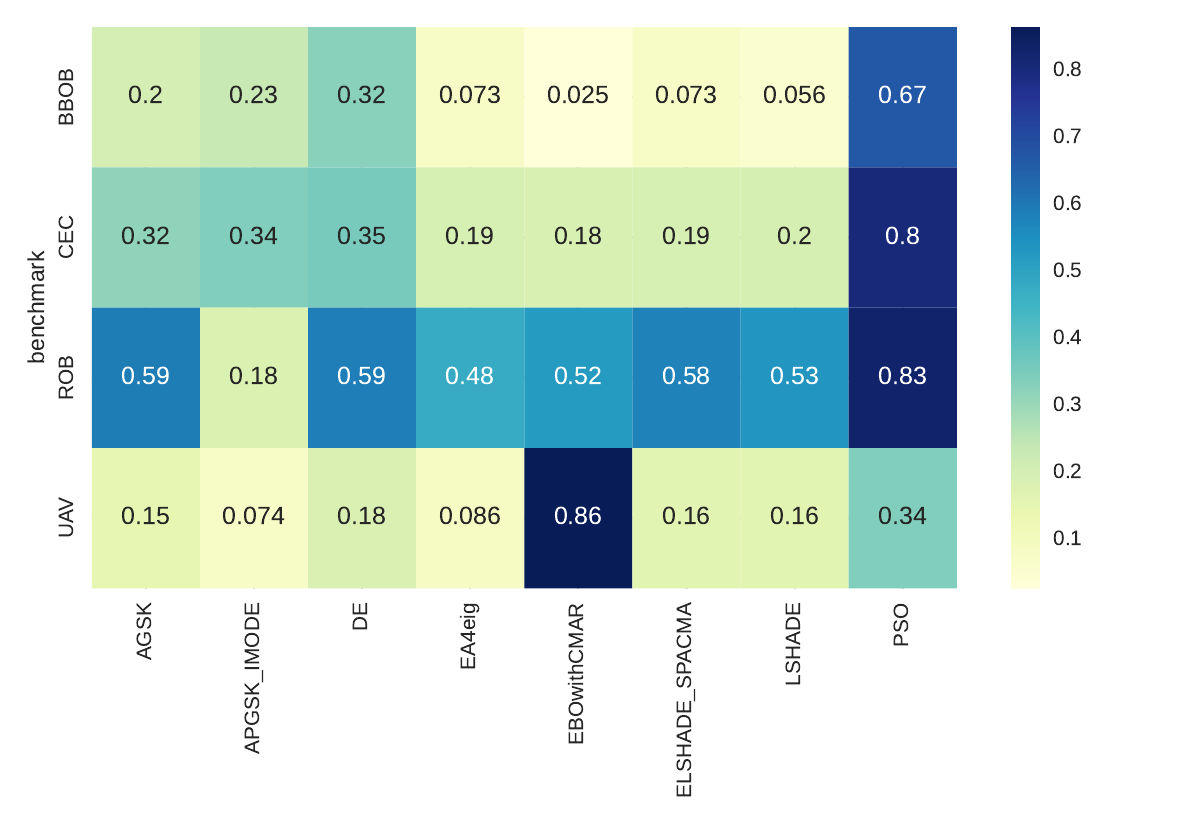}
    \caption{Mean algorithm performance per benchmark for 6$d$ problems}
    \label{fig:alg_perf_dim_6}
\end{figure}

\begin{figure}
    \centering
    \includegraphics[width=\linewidth]{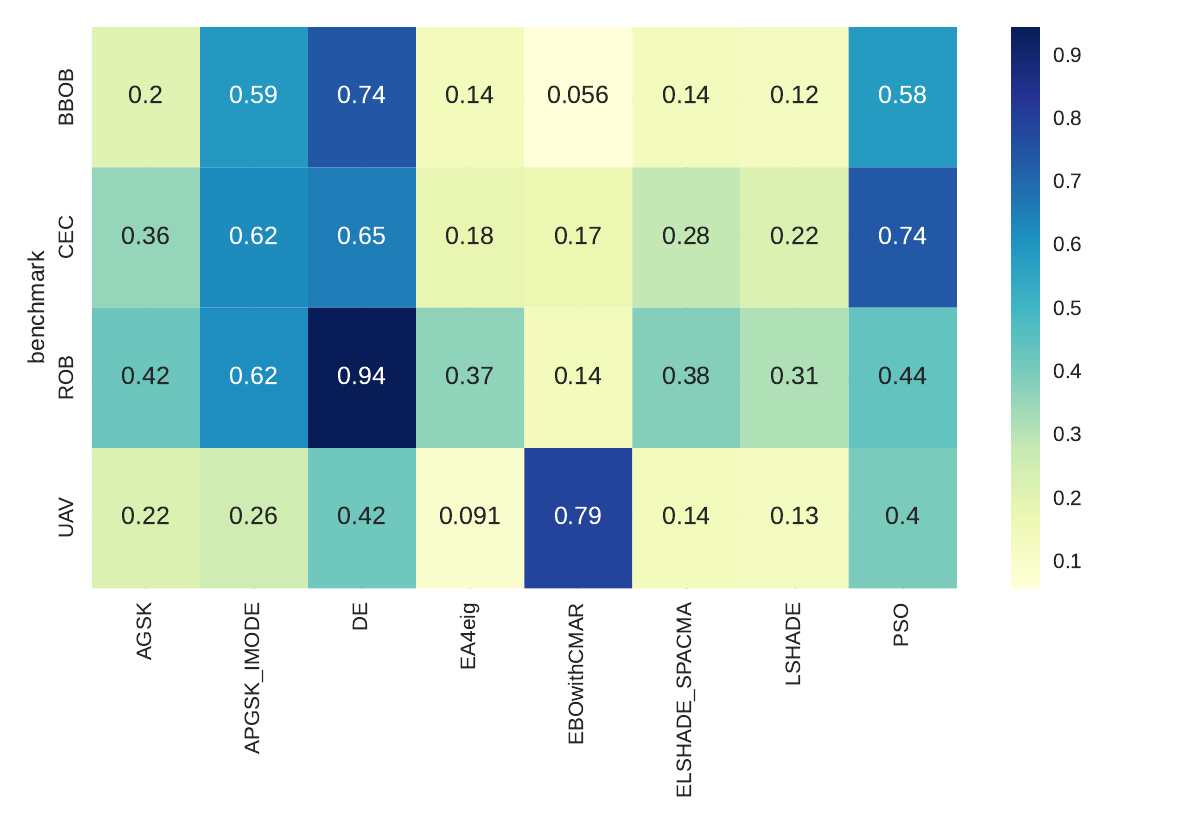}
    \caption{Mean algorithm performance per benchmark for 12$d$ problems}
    \label{fig:alg_perf_dim_12}
\end{figure}

\begin{figure}
    \centering
    \includegraphics[width=\linewidth]{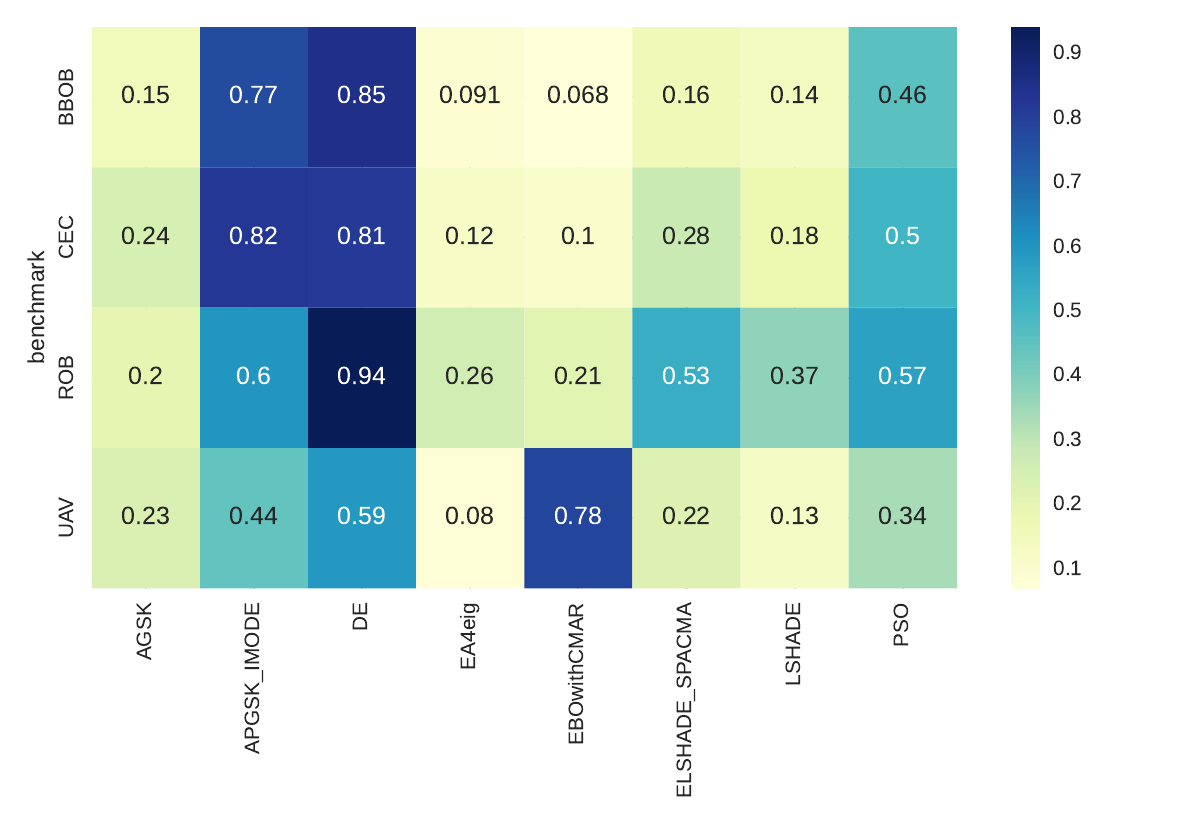}
    \caption{Mean algorithm performance per benchmark for 18$d$ problems}
    \label{fig:alg_perf_dim_18}
\end{figure}

\begin{figure}
    \centering
    \includegraphics[width=\linewidth]{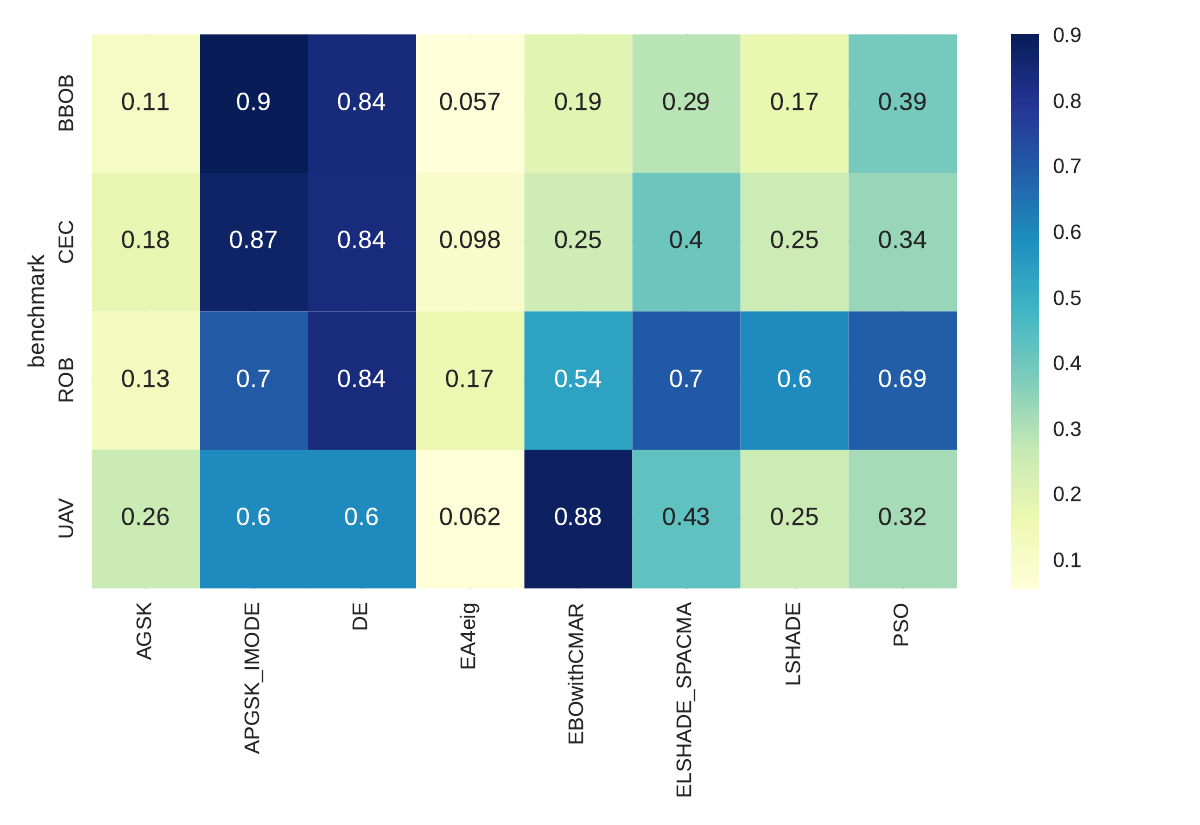}
    \caption{Mean algorithm performance per benchmark for 30$d$ problems}
    \label{fig:alg_perf_dim_30}
\end{figure}

Figures~\ref{fig:alg_perf_dim_6}-~\ref{fig:alg_perf_dim_30} feature the mean algorithm performance per benchmark, for each of the problem dimensions (lower is better). The figure demonstrating the results for 24$d$ problems can be found in the Supplementary Materials due to space limitations. We observe consistent yet benchmark-dependent patterns in algorithm performance as problem dimensionality increases. The four benchmarks (BBOB, CEC, ROB, and UAV) exhibit distinct algorithmic behaviors, revealing that no single method performs best across all settings. Overall, \texttt{EA4eig}, \texttt{EBOwithCMAR}, \texttt{ELSHADE\_SPACMA}, and \texttt{LSHADE} achieve the lowest mean normalized precision values, while \texttt{PSO}, \texttt{DE}, and \texttt{APGSK\_IMODE} typically rank among the weakest performers.

\begin{comment}
    \begin{figure}
    \centering
    \includegraphics[width=\linewidth]{figures/performance/algorithm_raw_performance_mean_per_benchmark_dim_24.pdf}
    \caption{Mean algorithm performance per benchmark for 24$d$ problems}
    \label{fig:alg_perf_dim_24}
\end{figure}
\end{comment}

The algorithm scores on the BBOB and CEC benchmarks are relatively consistent across dimensions. We typically have one algorithm that makes a strong single-best solver, and 4-5 of the other algorithms providing competitive performance, with scores under 0.2.
On 6$d$-18$d$ problems, the best-performing algorithm is \texttt{EBOwithCMAR}, with \texttt{EA4eig}, \texttt{ELSHADE\_SPACMA}, and \texttt{LSHADE} providing similar mean results. On 24$d$ and 30$d$ problems, the \texttt{EA4eig} algorithm provides the best results.

In contrast, results on the ROB benchmark vary considerably with problem dimensionality. On 6$d$ problems, \texttt{APGSK\_IMODE} achieves the lowest score (0.18), far ahead of the next-best algorithm (0.48). On 12$d$ problems, the performance of \texttt{APGSK\_IMODE} and \texttt{DE} worsens by 0.44 and 0.35, respectively, while the performance of \texttt{PSO} improves by 0.39, making \texttt{EBOwithCMAR} the top performer. On 18$d$-30$d$ problems, \texttt{AGSK} consistently ranks best.

On the 6$d$ UAV problems, the \texttt{APGSK\_IMODE} provides the best results. \texttt{EA4eig} is second ranked with only a 0.012 difference in mean performance. At higher dimensions, \texttt{EA4eig} consistently delivers the top performance.

\subsection{Algorithm Selection Results}

In this subsection, we present the results of the algorithm selection.

\begin{figure}
    \centering
    \includegraphics[width=\linewidth]{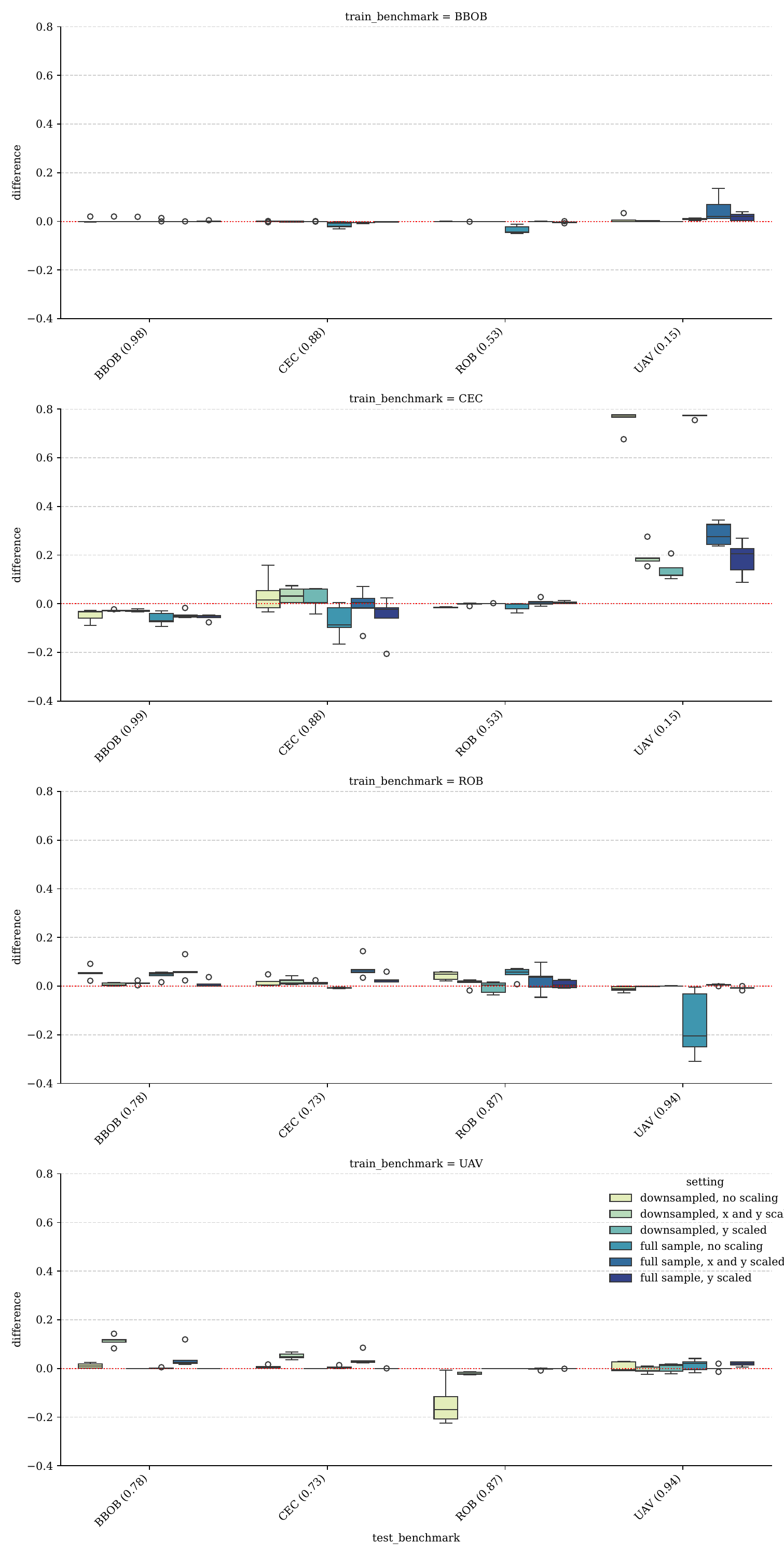}
    \caption{Difference between the performance of the RF model and the dummy model for each combination of training and testing benchmarks for 6$d$ problems}
    \label{fig:AS_results_dim_6}
\end{figure}

\begin{figure}
    \centering
    \includegraphics[width=\linewidth]{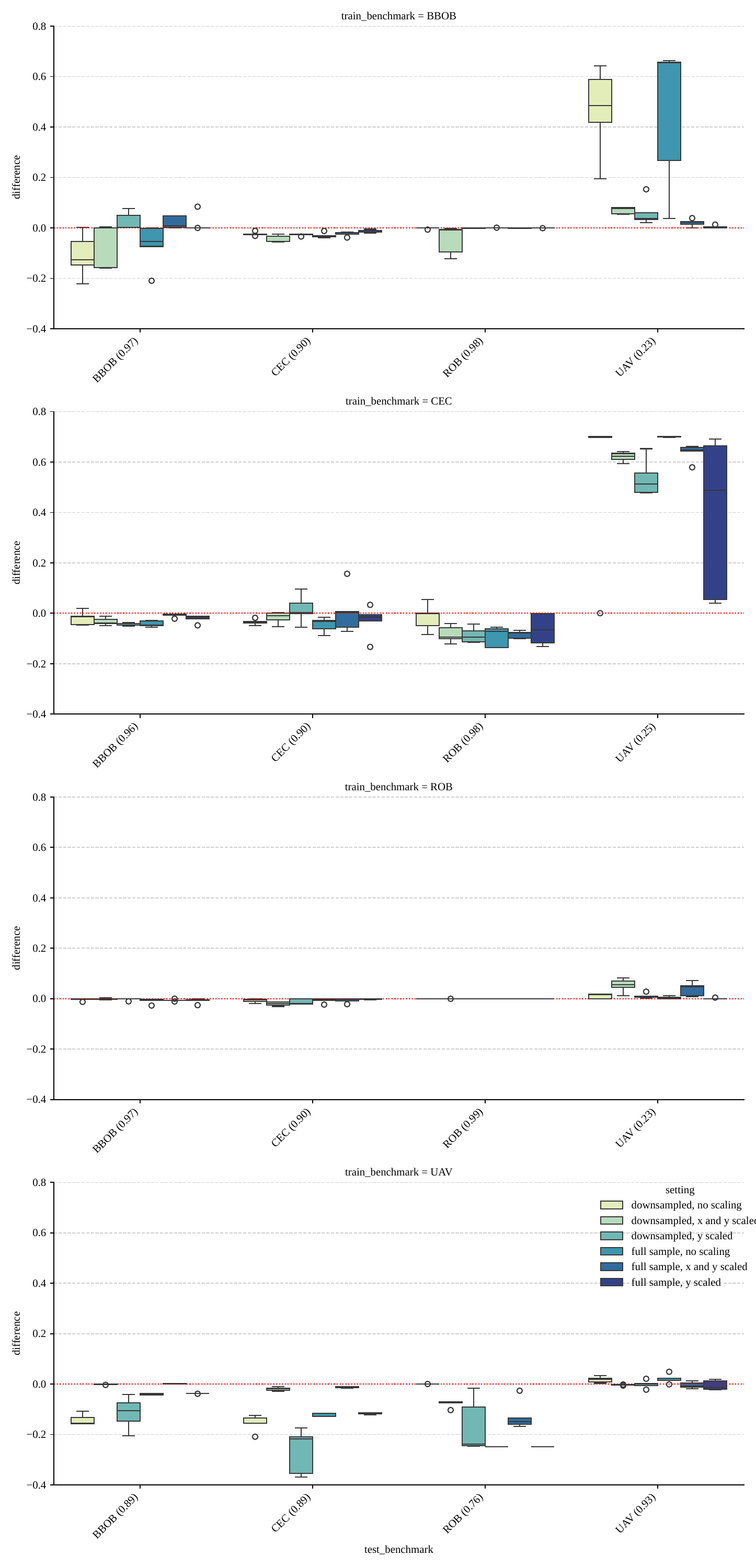}
    \caption{Difference between the performance of the RF model and the dummy model for each combination of training and testing benchmarks for 12$d$ problems}
    \label{fig:AS_results_dim_12}
\end{figure}

\begin{figure}
    \centering
    \includegraphics[width=\linewidth]{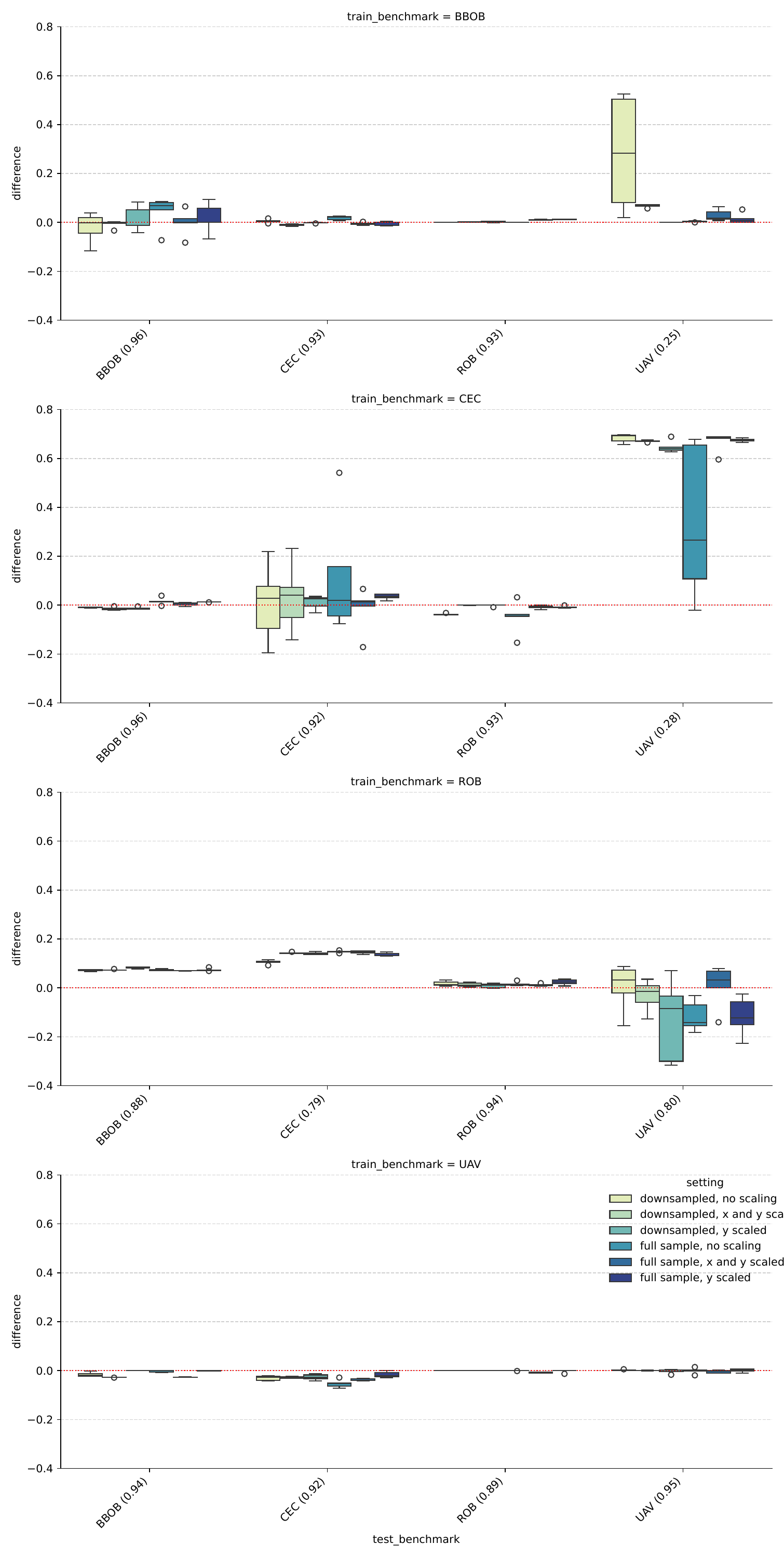}
    \caption{Difference between the performance of the RF model and the dummy model for each combination of training and testing benchmarks for 18$d$ problems}
    \label{fig:AS_results_dim_18}
\end{figure}

\begin{figure}
    \centering
    \includegraphics[width=\linewidth]{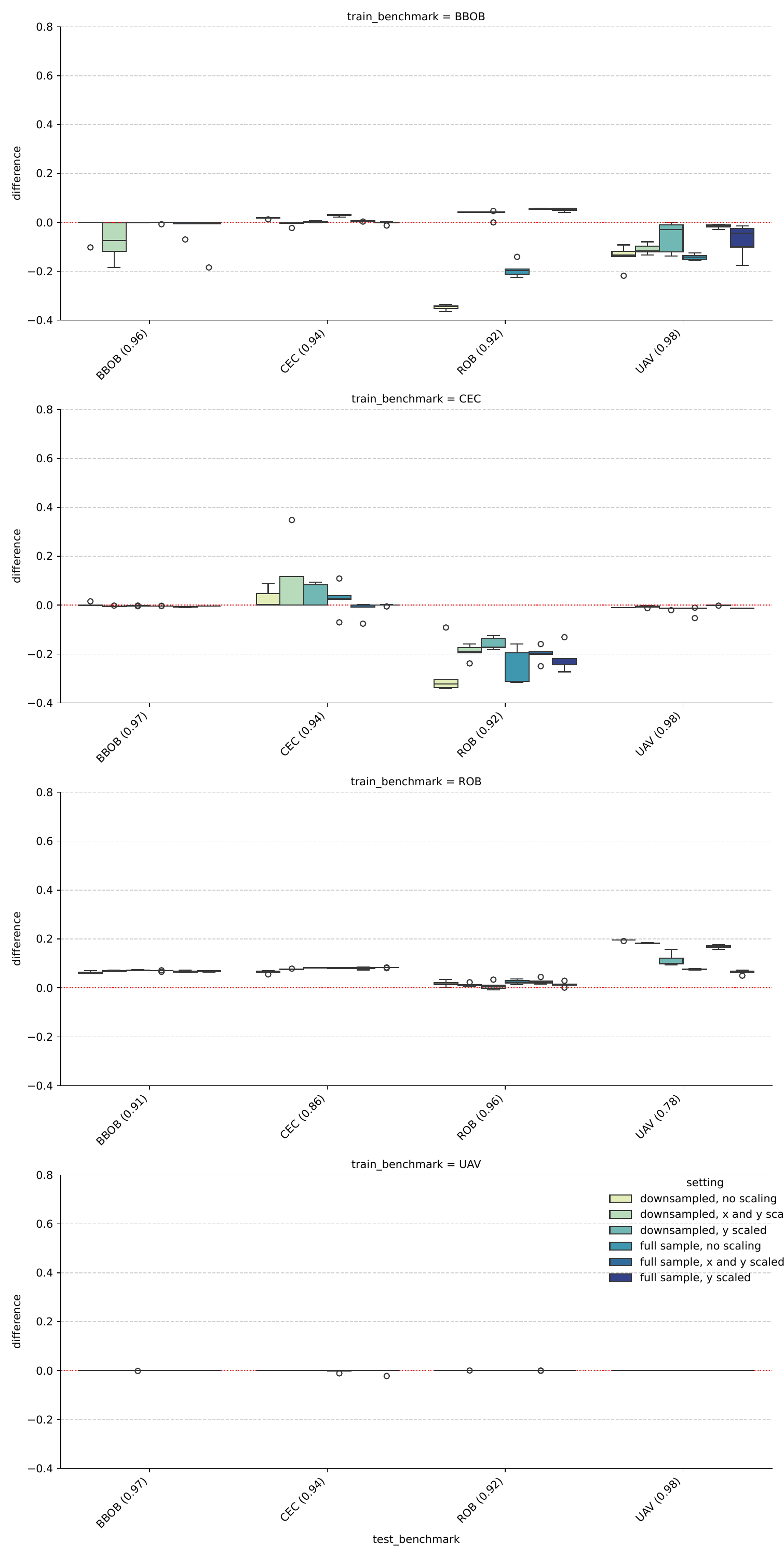}
    \caption{Difference between the performance of the RF model and the dummy model for each combination of training and testing benchmarks for 30$d$ problems}
    \label{fig:AS_results_dim_30}
\end{figure}

Figures~\ref{fig:AS_results_dim_6}--\ref{fig:AS_results_dim_30} present the differences in performance between the RF-based AS model and the dummy baseline across all combinations of training and testing benchmarks for dimensions ranging from 6$d$ to 30$d$. The figure demonstrating the results for 24$d$ problems can be found in the SM due to space limitations, however, results are most similar to 30$d$ problems. Each subplot corresponds to one training benchmark (BBOB, CEC, ROB, or UAV), with the boxplots showing the distribution of performance differences across test benchmarks depicted on the horizontal axis. Please note that we plot the difference between the RF model and the dummy so we can clearly see where the RF model outperforms the dummy, but we include the actual performance of the dummy in the brackets next to the name of the test benchmark on the horizontal axis.

\begin{comment}
\begin{figure}
    \centering
    \includegraphics[width=\linewidth]{figures/AS_results/AS_results_dim_24_stratify_True_difference.pdf}
    \caption{Difference between the performance of the RF model and the dummy model for each combination of training and testing benchmarks for 24$d$ problems}
    \label{fig:AS_results_dim_24}
\end{figure}
\end{comment}

\begin{figure*}
    \centering
    \includegraphics[width=0.8\linewidth]{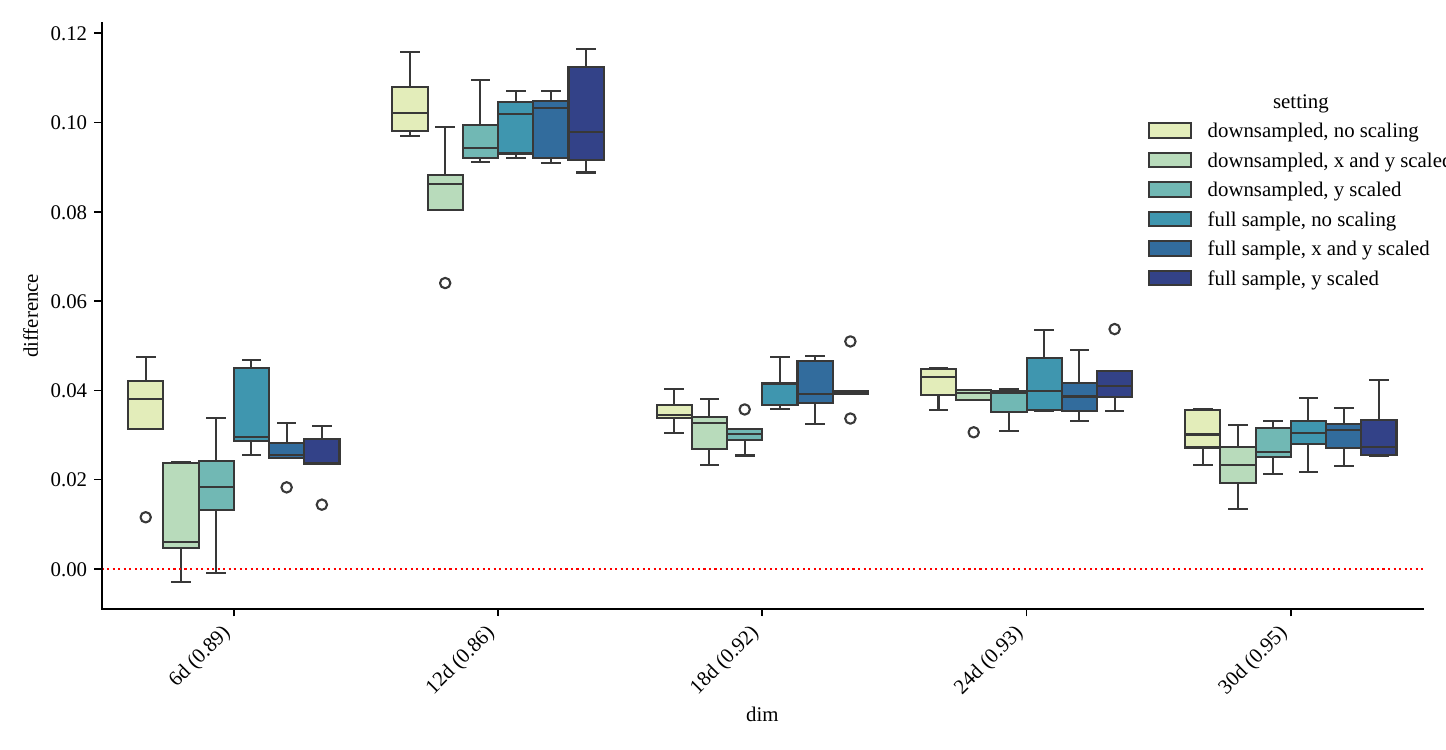}
    \caption{Difference between the performance of the RF model and the dummy model in the setting where benchmarks are merged together}
    \label{fig:AS_results_merged}
\end{figure*}

For the 6$d$ problems (Figure~\ref{fig:AS_results_dim_6}), we can see that in most cases, there is no generalization across the benchmarks. One exception is when the model is trained on the CEC benchmark and applied on UAV, where we achieve improvements as high as 0.8 over the dummy model when no scaling of x-values or y-values is applied. Interestingly, we can see that when scaling is applied, the improvement over the dummy is substantially lower, ranging from 0.1 to 0.4. This indicates that scaling may remove or distort informative absolute-scale characteristics that are important for cross-benchmark generalization in this setting.
We also observe that the RF model achieves marginal improvements over the dummy model when training is done on the ROB or UAV benchmark, and tested on the BBOB and CEC benchmarks.

On 12$d$ problems, we observe a successful transfer from the BBOB to the UAV benchmark when no scaling is performed. We also observe a successful transfer from CEC to UAV using all investigated features. 

On 18$d$ problems, we observe a successful transfer from the CEC to UAV, as well as from ROB to BBOB and CEC, also from BBOB to UAV when no scaling is applied, and downsampling is applied. This suggests that reducing the sampling density can act as a form of implicit regularization: by focusing on coarser, global landscape properties rather than fine-grained details, the resulting ELA features may become less sensitive to benchmark-specific artifacts and more suitable for cross-benchmark generalization.

On 24$d$ and 30$d$ problems, the AS trained on the ROB benchmark outperforms the baseline by about 0.1-0.2. Slight improvements over the baseline are also observed when transferring from the BBOB to the ROB benchmark, when using downsampling and scaling.

Figure~\ref{fig:AS_results_merged} shows the results from the experiment where all benchmarks are merged together, so the train and test sets can contain problems from any of the benchmarks. In this case, we can see that all AS models are better than the baseline and there is an improvement. This result provides a direction for future work, indicating that training models on data from all benchmarks is more effective for generalization and transfer to real-world benchmarks. Consequently, the training of AS models should shift toward an active learning setting.

\begin{figure}
    \centering
    \includegraphics[width=\linewidth]{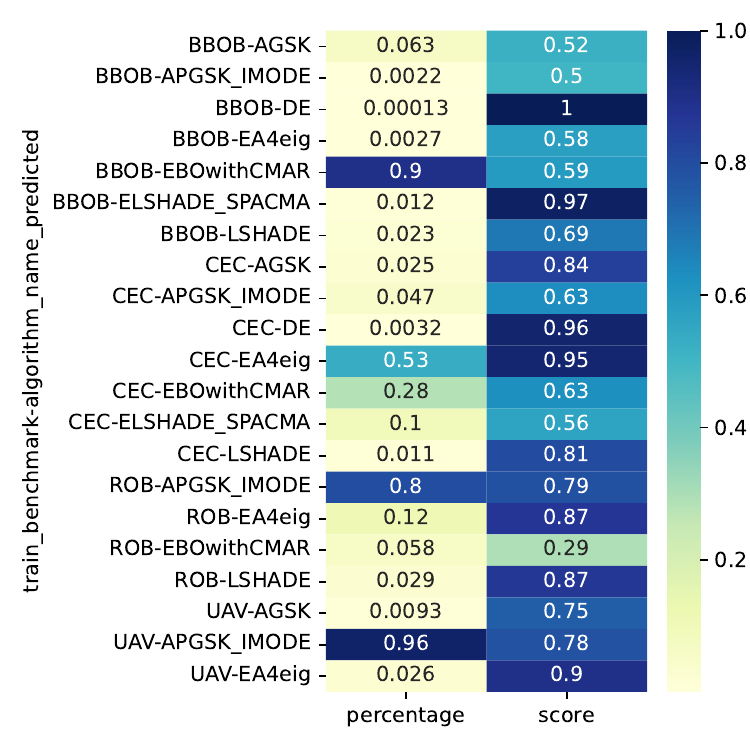}
    \caption{Predictions, 6$d$ problems}
    \label{fig:predictions_dim_6}
\end{figure}

Figure~\ref{fig:predictions_dim_6} shows, for each benchmark used to train the AS model, the percentage of times a given algorithm was predicted to be the best-performing method, together with the average prediction score associated with that selection. Each row corresponds to a single \textit{(training benchmark, algorithm)} pair. The first value represents the empirical frequency with which the algorithm is selected as the best, whereas the second value denotes the average quality of that prediction. Large percentages indicate that the meta-model consistently favors the algorithm when trained on that benchmark, whereas small percentages reflect rare selection. We can observe that models trained on benchmarks which have a strong single best solver typically just learn to favor that algorithm over all the others. We can see this happening when the model is trained on the BBOB, where the \texttt{EBOwithCMAR} algorithm is selected 90\% of the time, as well as when the model is trained on the UAV benchmark, where the \texttt{APGSK\_IMODE} algorithm is selected 96\% of the time. This explains why we see that models trained on such benchmarks do not outperform the dummy, since they are essentially learning to predict the same algorithm as the dummy.

\begin{figure}
    \centering
    \includegraphics[width=\linewidth]{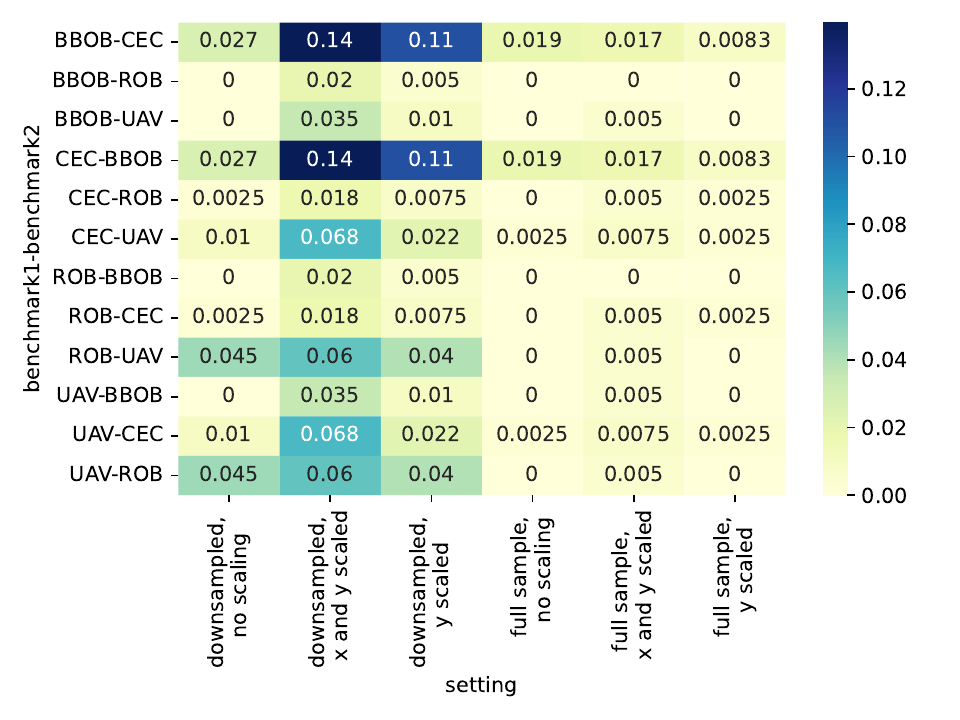}
    \caption{Benchmark similarity based on normalized crossmatch statistic, 6$d$ problems}
    \label{fig:crossmatch_dim_6}
\end{figure}

To further understand the lack of generalization, we investigated the feature-based similarity of the problems within the benchmarks. For this purpose, we employed the crossmatch test~\cite{crossmatch_test} for comparing multivariate distributions. 
Considering the ELA representations of problems within two benchmarks as two multivariate distributions, we use the crossmatch test to provide a nonparametric and distribution-free way to quantify similarity between them. By pooling the feature vectors from both benchmarks into a single set and forming an optimal nonbipartite matching that minimizes total pairwise distances, the test counts how often matched pairs consist of points originating from different benchmarks. This count (the crossmatch statistic) captures how intermingled the two sets of ELA representations are: a high number of crossmatches suggests that the benchmarks occupy largely overlapping regions of feature space, whereas a low number indicates that their underlying ELA distributions differ. We normalize this count by the maximum number of crossmatches for a pair of benchmarks (which is the minimum of the number of samples in both benchmarks), and use it as an indicator of overall benchmark similarity.
Figure~\ref{fig:crossmatch_dim_6} displays the normalized crossmatch test statistic for all benchmark pair comparisons on 6$d$ problems. Each row corresponds to a benchmark pair (e.g., BBOB–CEC, CEC–ROB, UAV–ROB), while the columns represent the different preprocessing settings. The values are mostly between 0 and 0.06, indicating weak crossmatching and therefore relatively distinct distributions among the benchmarks in low-dimensional feature space. The highest values appear for the BBOB-CEC pair (up to 0.14), suggesting that these two benchmarks exhibit the strongest overlap under several settings, but additional benchmark pairs—such as CEC–UAV and UAV–ROB—also show moderate normalized statistics under certain preprocessing settings.
Analyzing the effect of the preprocessing steps, we can see that scaling and downsampling makes the problems from different benchmarks more intertwined. 

\begin{figure}
    \centering
    \includegraphics[width=\linewidth]{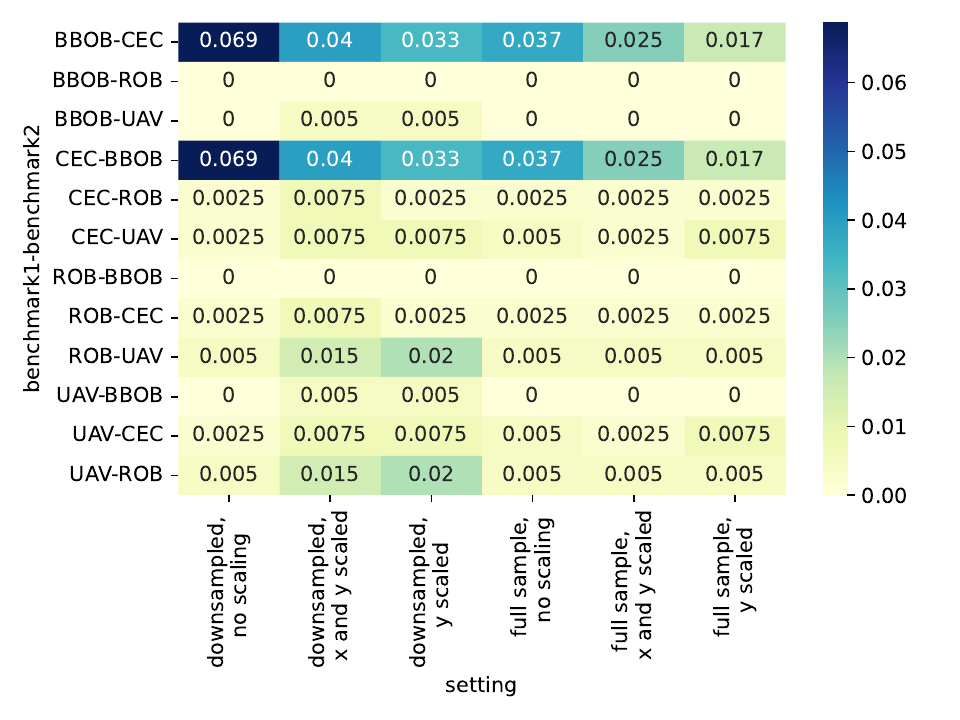}
    \caption{Benchmark similarity based on normalized crossmatch statistic, 30$d$ problems}
    \label{fig:crossmatch_dim_30}
\end{figure}

Figure~\ref{fig:crossmatch_dim_30} shows the same normalized crossmatch statistics but computed on 30$d$ problems. The values are even lower than for the 6$d$ problems, indicating even more distinct problems. The clearest pattern is again the high similarity between BBOB and CEC (up to 0.69), while many other pairs (especially those involving ROB or UAV) show near-zero crossmatch values, indicating very little similarity.

This analysis suggests that the real-world problems in the UAV and ROB benchmarks are substantially different than academic benchmark suites such as the CEC and BBOB.

%Figure~\ref{fig:top_features_dim_12} shows the feature importance values obtained from the RF model for the 10 features with the highest mean importance across all benchmarks for a 12$d$ problems. We analyze how the feature importance changes with different preprocessing techniques applied, indicated with the different colors of the barplots.
%\begin{figure*}
   % \centering
    %\includegraphics[width=\linewidth]{figures/top_features/top_features_dim_12.pdf}
    %\caption{Most important features, 12$d$ problems}
    %\label{fig:top_features_dim_12}
%\end{figure*}

%For the CEC benchmark, \texttt{ela\_meta.quad\_w\_interact.adj\_r2} becomes more important when scaling and downsampling are applied because scaling suppresses magnitude-related noise and highlights the smooth quadratic-like structure characteristic of many CEC functions. Downsampling reduces the reliability of ruggedness and nearest-neighbor features, causing the model to rely more heavily on global curvature signals. Since CEC landscapes are well approximated by quadratic models with interaction terms, the adjusted R2 coefficient of this model becomes a strong and stable predictor, increasing its importance.

\begin{comment}
\begin{figure*}
    \centering
    \includegraphics[width=\linewidth]{figures/top_features/top_features_dim_6.pdf}
    \caption{Most important features, 6$d$ problems}
    \label{fig:top_features_dim_6}
\end{figure*}
\end{comment}

\section{Conclusions}
\label{sec:conclusion}
This study examined the generalization ability of algorithm selection (AS) models beyond controlled synthetic benchmarks by systematically evaluating transfer across synthetic (BBOB, CEC) and real-world (robotics trajectory optimization and UAV path planning) problem suites. Our results show that, although benchmark suites such as BBOB and CEC remain indispensable for method development and controlled comparisons, AS models trained exclusively on these benchmarks often struggle to generalize to realistic, domain-specific problems.

Across dimensions, successful transfer was sporadic and strongly dependent on the source–target benchmark pair, the problem dimensionality, and preprocessing choices. In low dimensions (6$d$), generalization was largely absent, with only isolated cases of positive transfer (e.g., CEC to UAV without scaling). Moderate improvements emerged in some cases in higher dimensions. Nevertheless, these gains were typically modest, and in many cases AS models failed to outperform the dummy baseline.

The analysis of prediction frequencies revealed an important underlying issue: when a benchmark exhibits a dominant single best solver, AS models tend to overfit by consistently predicting that solver, effectively mimicking the dummy strategy. Such behavior explains the lack of improvement when transferring from benchmarks like BBOB or UAV, where one algorithm strongly dominates performance.

Our feature-based similarity analysis using the crossmatch test provides further insight into these findings. The generally low normalized crossmatch statistics indicate that synthetic and real-world benchmarks occupy largely distinct regions of the ELA feature space, especially in higher dimensions. While BBOB and CEC show substantial overlap, benchmarks involving robotics or UAV problems exhibit minimal similarity, which fundamentally limits cross-benchmark generalization. Preprocessing steps such as scaling and downsampling can increase feature-space overlap, but this does not consistently translate into stronger AS performance.

Overall, these results highlight a critical challenge for practical algorithm selection: strong performance on standard synthetic benchmarks does not guarantee robustness or transferability to real-world optimization problems. Future work should therefore focus on enriching benchmark collections with more diverse and realistic problem instances, developing representations that better capture transferable problem characteristics, and designing AS models explicitly aimed at cross-domain generalization rather than benchmark-specific performance.
%\begin{acks}
%The authors acknowledge financial support of the Slovenian Research Agency through program grant P2-0098, project grants J2-4460 and GC-0001. This work is also funded by the European Union under Grant Agreement 101187010 (HE ERA Chair AutoLearn-SI).
%\end{acks}

\bibliographystyle{ACM-Reference-Format}
\bibliography{references}

\appendix

% The authors acknowledge financial support of the Slovenian Research Agency through program grant P2-0098, project grants J2-4460 and GC-0001, and young researcher grant No. PR-12393 to GC. This work is also funded by the European Union under Grant Agreement 101187010 (HE ERA Chair AutoLearn-SI). 

\end{document}